# SDA-UCT: Self-Supervised Domain-Adaptive Ultrasound Computed Tomography for Rapid Musculoskeletal Sound Speed Reconstruction

Tianyu Liu*, Heyu Ma*, Aiduo Wang, Peiwen Li, Boyi Li, Ying Li, Dan Li, Chengcheng Liu, Dean Ta

***Abstract*—Ultrasound computed tomography (UCT) via full waveform inversion (FWI) enables high-resolution quantitative parametric imaging for biological tissue characterization and disease diagnosis. However, UCT iteratively reconstruct tissue acoustic properties (i.e., speed of sound (SOS)) based on a highly nonlinear optimization constrained by wave equation, suffering from large computational burden and severe convergence issues. Deep learning approaches can accelerate the UCT reconstruction, but the supervised training requires large-scale labeled datasets that are difficult to obtain for the *in-vivo* scenarios. Considering these limitations, we propose SDA-UCT, a two-stage self-supervised domain-adaptive framework for rapid and accurate UCT imaging of musculoskeletal tissues. The SDA-UCT first employ an attention-enhanced network (AttUCT) pre-trained on large-scale simulation datasets and then transfer to *in-vivo* datasets via physics-informed self-supervised learning, effectively bridging the simulation-to-real domain gap. Additionally, a Low-Rank Adaptation (LoRA) mechanism is integrated to enable efficient adaptation and improve the generalization capability across diverse clinical scenarios. Results showed that the proposed AttUCT achieved high-quality SOS reconstruction for simulated human forearm with a PSNR of 29.23 dB and SSIM of 0.928, outperforming conventional FWI and existing deep learning methods. Validated on *in-vivo* experimental data, the SDA-UCT successfully reconstructed the SOS images showing complex anatomical structures (including the skin, fat, muscle, tendon, bone and bone marrow, etc.) for human forearm, in high concordance with the MRI references. For special samples adaption, the LoRA mechanism adjusting only 3% of the parameters achieved comparable performance to the full fine-tuning. The rapid reconstruction time (i.e., only 5 ms per frame) enables SDA-UCT displaying the real-time three-dimensional (3D) visualization of tissue structures, achieving a five-orders-of-magnitude improvement over traditional FWI methods. This work represents the first self-supervised domain-adaptive deep learning for rapid and high-resolution UCT imaging *in vivo*, showing potential for disease diagnosis in musculoskeletal tissues.**

## I. Introduction

Quantitative parametric imaging of biological tissues provides essential biomarkers for disease diagnosis and treatment monitoring. Conventional imaging modalities have inherent limitations: MRI and CT primarily visualize structural features rather than acoustic properties [1, 2], while B-mode ultrasound suffers from poor spatial resolution, limited contrast, and operator dependency [1, 2]. These limitations motivate the development of quantitative acoustic imaging techniques for tissue characterization.

Recently, ultrasound computed tomography (UCT) via full waveform inversion (FWI) enables high-resolution quantitative imaging of the acoustic parameters (e.g., speed of sound (SOS), attenuation coefficient, and density) for biological tissue [3-6]. Unlike conventional beamforming methods that approximate wave propagation as straight rays [7, 8], FWI solves a nonlinear optimization problem constrained by the complete wave equation [9]. FWI iteratively minimizes the misfit between the measured and simulated waveforms by updating the acoustic parameters (e.g., SOS, density) which are involved in the wave equation [4, 9, 10]. Reconstructing standard UCT imaging with high interpretability, this physics-based FWI approach has demonstrated promising performance in soft tissue characterization [11-13]. Clinical UCT has matured primarily in breast imaging applications, including density assessment, cancer detection, and treatment monitoring [11, 14, 15]. However, UCT iteratively reconstruct tissue acoustic properties based on a highly nonlinear optimization constrained by wave equation, suffering from large computational burden and severe convergence issues.

There are many technical challenges obstructing UCT from a wide clinical application in the diagnosis of various biological

This work was supported by the National Natural Science Foundation of China (W2511001, 82127803, 12327807) and National Key R&D Program of China (2023YFC2410800). The computations in this research were performed using the CFFF platform of Fudan University. *(Corresponding author: Chengcheng Liu, Dean Ta).*

Tianyu Liu, Heyu Ma, Peiwen Li, Aiduo Wang, Boyi Li, Ying Li are with the College of Biomedical Engineering, Fudan University, Shanghai 200433, China (email: 23110860039@m.fudan.edu.cn; 22110860029@m.fudan.edu.cn; 23110720098@m.fudan.edu.cn; 23110860030@m.fudan.edu.cn; liboyi@fudan.edu.cn; yl@fudan.edu.cn).

Dan Li is with the College of Future Information Technology, Fudan University, Shanghai 200433, China (email: lidan@fudan.edu.cn).

Chengcheng Liu and Dean Ta are with the College of Biomedical Engineering, Fudan University, Shanghai 200433, China, and also with State Key Laboratory of Integrated Chips and Systems, Fudan University, Shanghai 201203, China (email: chengchengliu@fudan.edu.cn; tda@fudan.edu.cn).

*These authors contributed equally to this work.

tissues. First of all, FWI is an ill-posed inversion and could easily be trapped in local minimum with the so-called 'cycle skipping' phenomenon, especially for imaging the musculoskeletal or cranial tissues with large acoustic impedance discrepancy [16]. The large mismatch in SOS between soft tissue (~1540 m/s [17] ) and bone (~2900 m/s [18]) leads the inverse problem highly nonlinear with multiple local minima, frequently causing the convergence to suboptimal solutions [19]. Therefore, the UCT imaging for musculoskeletal or cranial tissues are degraded to low quality, typically with numerous ring artifacts. Another big problem for UCT is the computational burden and prohibitively long reconstruction time associated with the iterative inversions. The iterative optimization process requires repeated forward simulations for wave propagation, resulting in reconstruction times ranging from tens of minutes to several hours per case [4, 9, 20]. Such a long reconstruction time is not acceptable and incompatible with the time-sensitive clinical applications such as intraoperative guidance and dynamic functional imaging.

Recent advances in deep learning have introduced promising alternatives to FWI by directly learning the inverse mapping from ultrasound waveforms to SOS distributions [19, 21, 22]. Representative methods include cascaded encoder-decoder networks for near real-time reconstruction [23], dual-encoder architectures with high-frequency enhancement for bone imaging [24], and CNN-based transcranial aberration correction [11]. These data-driven approaches bypass the iterative optimization through supervised training on large-scale simulation datasets. The deep learning frameworks have significantly accelerated the inversion and achieved millisecond-level reconstruction times, which are several orders of magnitude faster than conventional FWI-based UCT.

Despite the computational advantages, deep learning-based reconstruction faces several critical limitations, especially when considering for the clinical translation. First, these supervised learning methods requires accurate SOS ground truth, which is prohibitively difficult to obtain in clinical scenarios. Actually, there are no such kinds of techniques that can provide high-accuracy spatially-distributed SOS standards for biological tissues *in vivo*. The gold-standard imaging modality like MRI displays tissue structure associated with the volume content of nucleus (i.e., mainly the 1H protons), not acoustic properties like SOS [23, 24]. Second, simulation-to-real domain gap would significantly degrade the imaging performance when the simulation-trained models transfer to the *in vivo* applications. This domain gap is mainly due to the mismatches in the SOS distribution as well as the waveforms between the simulated and experimental data [24, 25]. Third, existing deep learning methods primarily focusing on the imaging for soft tissue with relatively uniform SOS variations [10, 26], while their effectiveness for musculoskeletal tissues with highly discrepant SOS remains uncertain, as complex wave phenomena pose substantially greater reconstruction challenges. Fourth concern is about the generalization capability of the deep learning frameworks. Currently, the pre-trained supervised models assume relatively consistent or normal scenarios for the datasets. The imaging performance would significantly degrade and fail to account for atypical cases, limiting their generalization capability across diverse clinical scenarios.

Considering these aforementioned limitations, we proposed the SDA-UCT, a two-stage self-supervised domain-adaptive framework for rapid and accurate musculoskeletal sound speed reconstruction. Considering the lack of SOS ground truth *in vivo*, we developed a physics-informed self-supervised learning framework that exploits wave equation constraints as supervisory signals, enabling model training directly on the experimental data without the need for *in-vivo* ground truth SOS. Consequently, to bridge the simulation-to-real domain gap, SDA-UCT employs a two-stage domain-adaptive strategy: supervised pre-training on large-scale simulation datasets followed by physics-informed self-supervised fine-tuning on in-vivo experimental measurements, effectively transferring the learned representations to real clinical scenarios. To achieve a high reconstruction performance for tissues with large SOS discrepancy, we proposed the end-to-end attention-enhanced network architecture (named AttUCT) specially designed to capture the complex wave propagation characteristics in the heterogeneous musculoskeletal tissues. To improve the generalization capability, we integrated a Low-Rank Adaptation (LoRA) mechanism to achieve patient-specific fine-tuning with minimal parameter updates for atypical samples. The proposed SDA-UCT was validated on *in-vivo* human forearms, demonstrating a high-quality SOS imaging comparable to MRI references, but with fast millisecond-level reconstruction times. The main contributions of this study include:

(1) A physics-informed self-supervised learning framework exploiting wave propagation constraints that enables training without the need for *in-vivo* ground-truth SOS labels;

(2) A two-stage domain-adaptive strategy that combines the supervised pre-training on large-scale simulation datasets with the self-supervised fine-tuning on *in-vivo* experimental measurements, effectively bridging the simulation-to-real domain gap;

(3) An attention-enhanced network architecture (AttUCT) specifically designed for UCT imaging of musculoskeletal tissues with large SOS discrepancy;

(4) A LoRA-based adaptation mechanism that adjusts a small subset of trainable parameters while improving the generalization capability across atypical samples;

(5) The first in-vivo validation on human forearms achieving high-quality 3D UCT imaging with 5 ms reconstruction time per frame, five orders of magnitude faster than traditional FWI.

## II. Theory and Methods

### A. Forward Modeling for Ultrasonic Propagation

Ultrasonic wave propagation through heterogeneous biological tissues can be described by the acoustic wave equation, with the acoustic pressure field following:

$$\frac{1}{c(x,z)^2}\frac{\partial^2 P(x,z,t)}{\partial t^2} - \nabla^2 P(x,z,t) = S(x,z,t) \quad (1)$$

where $P(x,z,t)$ is the acoustic pressure field in two-dimensional region $(x,z)$, $c(x,z)$ denotes the spatially distributed SOS maps, and $S(x,z,t)$ represents the ultrasonic source excitation.

Figure 1. Architecture of the SDA-UCT framework. The two-stage strategy includes: (Stage 1) supervised pre-training on simulation data; (Stage 2) self-supervised fine-tuning on experimental data.

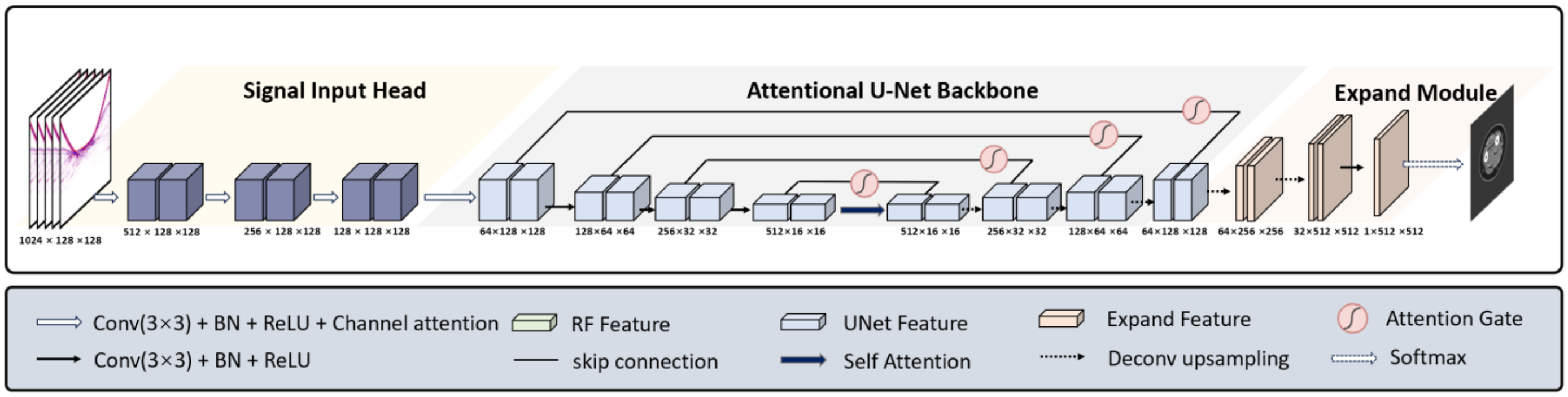

Figure 2. Network architecture of AttUCT with signal input head, attentional UNet backbone, and expand module.

In a UCT system, the transmitting transducers excite acoustic waves at source positions $e_s$, and the receiving transducers record ultrasonic pressure signals at receiver positions $e_r$. The ultrasonic signal recorded by each receiver can be expressed as:

$$d(e_r, t; c) = P(e_r, t; c, e_s) \tag{2}$$

where $d(e_r, t; c)$ represents the time-domain signal received at position $e_r$, which is a function of the SOS map $c$ and source positions $e_s$.

### B. Self-Supervised Domain-Adaptive UCT (SDA-UCT)

The proposed SDA-UCT framework employed the two-stage domain-adaptive strategy that combines the supervised attention-enhanced data-driven model (AttUCT) with physics-informed self-supervised training methodology, followed by a LoRA-based fine-tuning mechanism.

#### 1) Two-Stage Domain-Adaptative Training Strategy

As illustrated in Figure 1, the two-stage training strategy in SDA-UCT bridges the simulation-to-real domain gap through progressive knowledge transfer. In the first stage, supervised learning (AttUCT) was performed on large-scale simulation datasets, enabling the model to learn the direct mapping from radio-frequency (RF) signals to SOS distributions by minimizing the mean squared error between predicted and ground truth SOS distributions. In the second stage, the pre-trained AttUCT model was transferred to experimental scenario and fine-tuned using unlabeled experimental data through physics-informed self-supervised learning. Specifically, the network-predicted SOS maps were fed into a differentiable wave equation solver, integrated in the training pipeline to generate simulated RF signals. The physics-informed loss was computed by comparing the simulated and experimental RF signals, enabling gradient-based network optimization.

#### 2) Supervised and Self-Supervised Loss Functions

In the first-stage training on simulation datasets, fully supervised learning was employed with ground-truth SOS maps, using the following loss function:

$$L_{stage1} = \frac{1}{N}\sum_{i=1}^{N}\left\| c_i - c_i^{gt} \right\|_2^2 \tag{3}$$

where $\hat{c}_i$ represents the predicted SOS map, $c_i^{gt}$ denotes the ground-truth SOS map, and $N$ is the number of simulated training samples.

In the second-stage training on experimental measurements, physics-informed self-supervised learning was employed to address the absence of ground-truth SOS maps for experimental samples. Self-supervised learning exploits inherent physical constraints as supervision [27-32], eliminating the need for labeled data. In the second-stage of SDA-UCT, forward acoustic modeling was performed using the network-predicted SOS maps to generate simulated RF signals $d_{pre}$, which were subsequently compared with the experimentally acquired signals $d_{obs}$. The self-supervised loss function was formulated as:

$$L_{stage2} = \frac{1}{N_r} \sum_{r=1}^{N_r} \int_0^T \left[ d_{obs} - d_{pre}(e_r, t; c) \right]^2 dt \tag{4}$$

where $d_{pre}(e_r, t; \hat{c})$ represents the forward-simulated signal computed using Eq. (2) with predicted SOS map $\hat{c}$.

#### 3) Attention-Enhanced Data-Driven UCT Model (AttUCT)

To learn the mapping from raw RF signals to SOS maps for musculoskeletal tissues, we designed the end-to-end AttUCT architecture (Figure 2). The AttUCT network takes ultrasound RF matrix as input with dimensions [1024, 128, 128], representing temporal signal length, number of transmit events, and number of receiving transducers, respectively. The output of AttUCT is a reconstructed SOS map with dimensions [1, 512, 512]. The AttUCT model comprises three key components:

(1) Signal Input Head: A channel attention mechanism transforms raw RF signals from the temporal domain to compact feature representations, with attention weights computed along the temporal dimension to emphasize informative signal components.

(2) Attentional U-Net Backbone: An encoder-decoder architecture with gate mechanism at skip connections, and a self-attention module at the bottleneck for global spatial dependency modeling.

(3) Expand Module: Deconvolutional layers upsample features to the target SOS map resolution.

#### 4) LoRA Fine-Tuning Mechanism

While the two-stage domain-adaptive strategy in SDA-UCT achieves robust performance on typical samples, atypical samples with abnormal SOS distributions may exhibit suboptimal reconstruction quality. For such special samples, the model can be fine-tuned to achieve patient-specific adaptation. To facilitate continuous follow-up examinations, the fine-tuned weights can be stored for each individual patient. However, full-parameter storage poses significant storage and deployment challenges. To address this issue, we integrate Low-Rank Adaptation (LoRA) [33-35] into the SDA-UCT framework. LoRA exploits the low intrinsic dimensionality of parameter updates during model adaptation. For a weight matrix $W_0 \in \mathbb{R}^{d\times k}$ in the pre-trained model, LoRA represents parameter updates through low-rank decomposition matrices:

$$W = W_0 + \Delta W = W_0 + BA \tag{5}$$

where $A \in \mathbb{R}^{r\times k}$ and $B \in \mathbb{R}^{d\times r}$ are trainable low-rank matrices, and $r$ is the rank parameter. During fine-tuning, the original weights $W_0$ remain frozen, with only matrices $A$ and $B$ being updated. Since $r \ll \min(d, k)$, the number of updated parameters is significantly smaller than the total parameters in $W$. By applying LoRA to our framework, we can create a patient-specific adapter for each atypical case (Figure 3), achieving high-quality robust reconstruction with minimal storage overhead while preserving the shared pre-trained backbone.

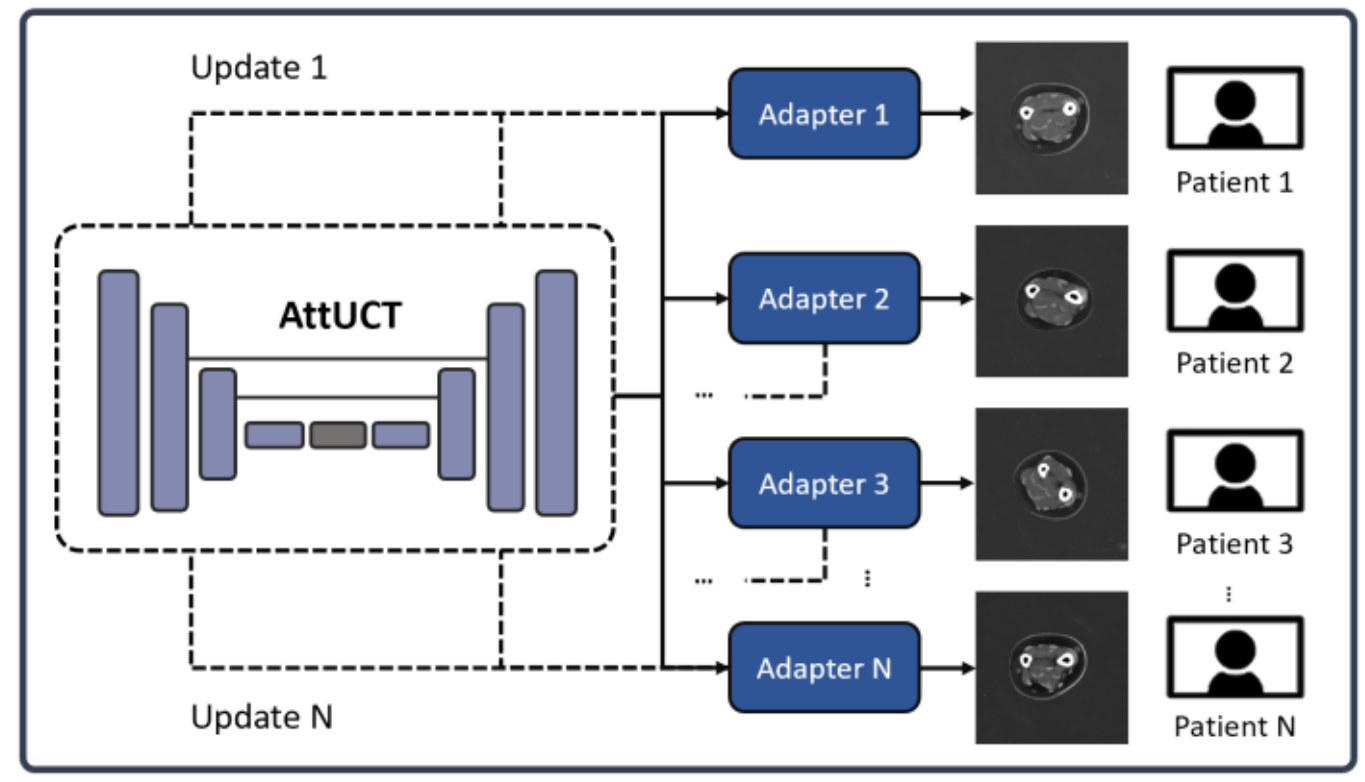

Figure 3. Patient-specific adaptation with LoRA for atypical samples in SDA-UCT.

## III. EXPERIMENTS AND RESULTS

### A. Datasets and Evaluation metrics

#### 1) Simulation Dataset

Figure 4 shows the workflow to construct MRI-based ground-truth SOS maps for simulation dataset. 10 healthy volunteers (5 males and 5 females, age range: 20–35 years) underwent the clinical high-field MRI (5 T, T2-mode, uMR Jupiter) for high-resolution MRI scans of human forearm. The MRI imaging protocol yielded a pixel resolution of 0.23 × 0.23 × 3.30 mm³, with 25 axial slices per volunteer and totally 250 slices for all volunteers (Fig.1(a) and (b)). In order to convert MRI images to SOS maps, the distinct anatomical regions in forearm (including bone, muscle, tendon, fat, skin, etc.) were delineated using a fine-tuned Segment Anything Model (SAM) for semi-automated segmentation [36] (Fig.1(c)). The anatomical regions for each type of tissue were assigned with the empirical SOS values (Table 1). The dataset was augmented fourfold through random rotation and flipping operations, generating a total of 1000 simulation datasets. Note that although named as ground-truth SOS maps for simulation, the assigned empirical SOS values were just a rough guess for these tissues and may be significantly different from the true SOS in the forearm, leading to the simulation-to-real domain gap.

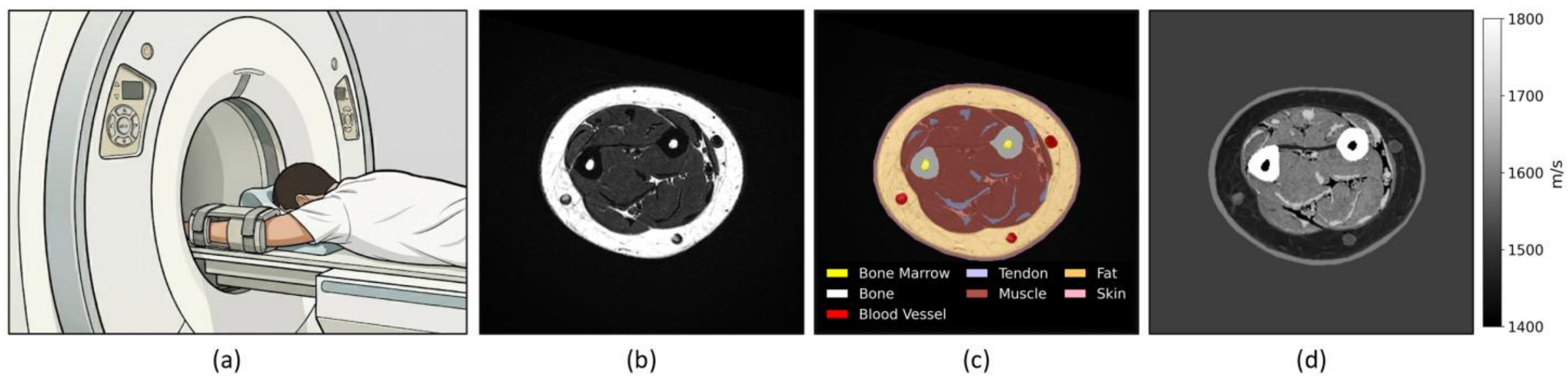

Figure 4. Workflow for constructing MRI-based ground-truth SOS maps: (a) clinical MRI scan of human forearm; (b) high-resolution MRI slice; (c) segmented anatomical regions; (d) ground-truth SOS map with assigned empirical values.

Ultrasound propagation in human forearm was simulated by solving the wave equation (1), with the assigned ground-truth SOS maps. As illustrated in Fig.1(d), the simulation employed a 128-element ring array transducer with a 100-mm diameter on a 512×512 computational grid with 0.2-mm spatial interval. Each element of the transducer transmitted ultrasonic excitation sequentially in turn and all 128 elements received ultrasonic signals simultaneously. For each SOS map, this kind of 128 transmission-reception cycles yielded ultrasonic tomographic full-matrix signals with the dimension of 1024 (signal time length) ×128 (transmitting number) ×128 (receiving number). Ultrasonic signal was excited with a 1 MHz center frequency source and sampled at 25 MHz. This simulation protocol generated 1000 SOS map-UCT signal pairs for the simulation datasets, with 800 pairs allocated for training and 200 pairs for testing.

Table 1. Empirical SOS values assigned for delineated tissues in human forearm

| Human tissues | SOS (m/s) |
| --- | --- |
| Bone | 2900 |
| Bone Marrow | 1450 |
| Blood | 1584 |
| Tendon | 1670 |
| Skin | 1580 |
| Fat | 1430-1478 |
| Muscle | 1540-1630 |

*2) Experimental Dataset*

As illustrated in Fig.1 (Stage 2), *in-vivo* ultrasonic tomographic measurements were performed on human forearm using a ring-array transducer (128 elements, inner diameter of 10 cm, a central frequency of 1 MHz) with the ultrasound system (Vantage 256, Verasonics Inc, Kirkland, WA). Consistent with the configuration in simulations, the *in-vivo* tomographic full-matrix signals for the second-stage training were also with the dimension of 1024 (signal time length) ×128 (transmitting number) ×128 (receiving number). The *in-vivo* experimental dataset was measured from 8 healthy volunteers (4 males, 4 females, aged 20–35 years). To ensure data independence and enable unbiased evaluation of domain adaptation, the 8 volunteers for the *in-vivo* experimental dataset were entirely different from the 10 participants for the simulation dataset. All other acquisition parameters were configured to match the simulation settings to maintain cross-domain consistency. 20 cross-sectional ultrasonic tomographic scans were acquired for each volunteer, yielding a total of 160 in-vivo experimental samples. Notably, these *in-vivo* experimental samples contained only raw RF signals, without corresponding accurate and real SOS maps.

Data from 7 volunteers (140 samples) were used for the second-stage self-supervised experiment and split into training and testing sets with an 8:2 ratio. The remaining volunteer, a female with intensive fitness habits (upper limb strength training ≥4 times/week), exhibited significantly higher muscle mass and distinct SOS distributions compared to typical female subjects. Her 20 samples were reserved for the LoRA fine-tuning experiment (16 for training and 4 for testing) to evaluate adaptation performance on special samples.

The study was approved by the Ethics Committee of Fudan University. Informed consent was obtained from all participants. All procedures involving human participants were conducted in accordance with the Declaration of Helsinki and its later amendments or comparable ethical standards.

*3) Evaluation Metrics*

Stage-specific evaluation strategies were employed to assess the performance of the SDA-UCT framework. For the first-stage supervised training for simulation dataset with known ground-truth SOS maps, the SOS reconstruction quality is quantified using three complementary metrics: Peak Signal-to-Noise Ratio (PSNR) for pixel-level accuracy [37], Structural Similarity Index Measure (SSIM) for structural fidelity [38], and Learned Perceptual Image Patch Similarity (LPIPS) for perceptual quality [39]. For the second-stage self-supervised training for *in-vivo* experimental dataset where the ground-truth SOS maps are unavailable, we evaluate the waveform consistency using L2 loss between the simulated signals with predicted SOS maps and the measured signals from human forearm *in vivo*.

## B. Implementation Details

For the first-stage model training on simulation dataset, we utilized 800 SOS map-signal data pairs with training conducted over 200 epochs using a batch size of 4. The learning rate was initialized at $1\times10^{-3}$ with cosine decay scheduling, and the AdamW optimizer was employed for parameter optimization.

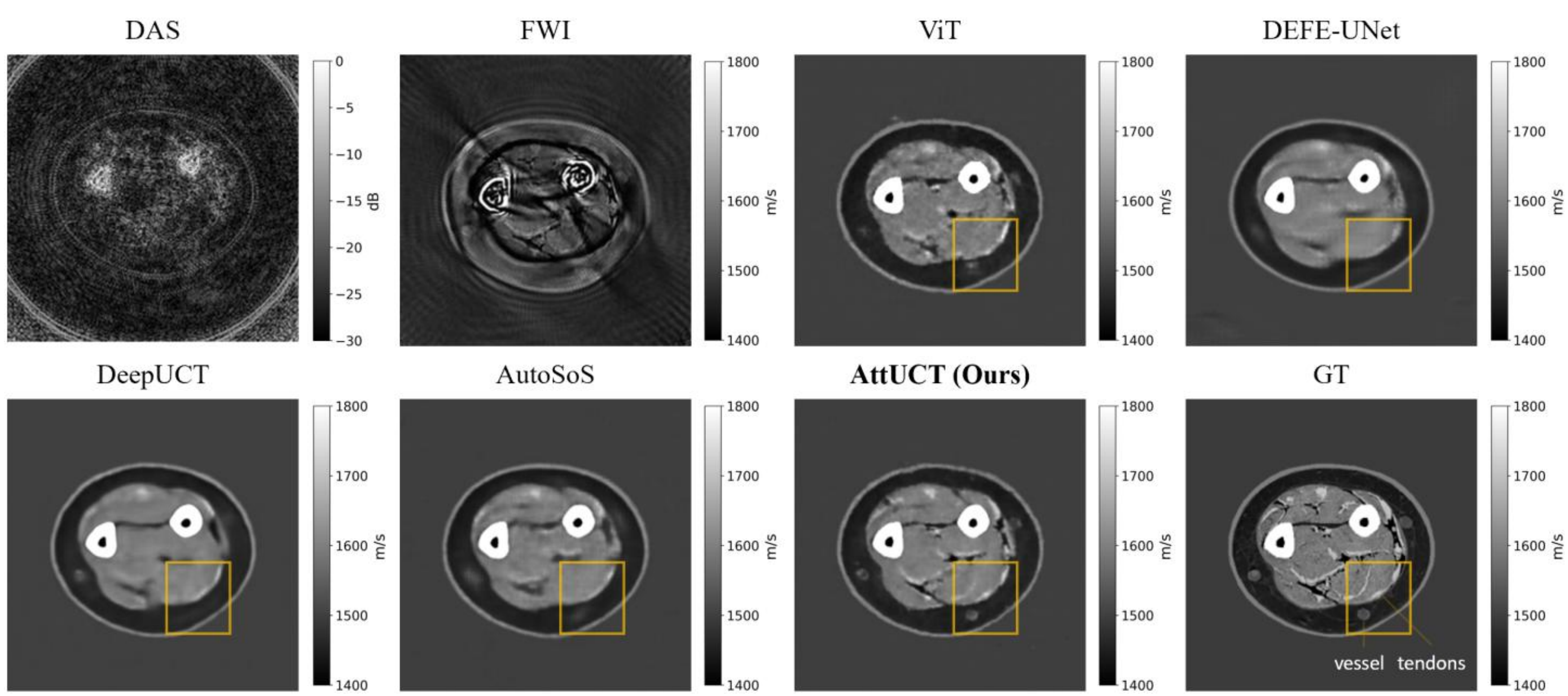

Figure 5. Comparison of supervised AttUCT with conventional and deep learning methods on simulated data. Yellow boxes highlight small tissue structures (vessels and tendons).

For the second-stage self-supervised training on in-vivo experimental dataset, full-parameter fine-tuning was performed using experimental data from 7 volunteers. All network parameters were updated over 100 epochs with a batch size of 1, utilizing a reduced learning rate of $1\times10^{-4}$ with cosine decay scheduling and the AdamW optimizer. The reduced batch size and learning rate were designed to enable careful adaptation to experimental data while preserving the learned representations from the simulation-based first stage.

For the LoRA fine-tuning experiments on the special volunteer, two adaptation strategies were compared: full-parameter fine-tuning and LoRA-based fine-tuning. The full-parameter fine-tuning followed the same configuration as the second stage. For LoRA-based fine-tuning, low-rank adaptation modules were integrated into the convolutional layers of AttUCT with rank $r = 16$ and scaling factor $\alpha = 16$, maintaining identical training configurations (60 epochs, batch size of 1, learning rate of $1\times10^{-4}$). All training was conducted on a single NVIDIA A100 GPU.

## C. Supervised Training on Simulated Datasets

Supervised training and testing were conducted on the simulation dataset with ground-truth SOS maps. The proposed AttUCT was compared against traditional DAS, FWI methods and mainstream deep learning methods, including AutoSOS [22], DeepUCT [23], DEUNet-FE [24], and Vision Transformer (ViT) [33].

### 1) Comparison with Traditional Methods

Figure 5 compares AttUCT against conventional DAS beamforming and FWI with L-BFGS optimizers. DAS produced severely blurred B-mode images showing only rough bone outlines due to acoustic impedance mismatch at tissue interfaces. FWI achieved improved structural visibility but suffered from substantial ring-shaped artifacts in bone regions due to cycle-skipping phenomena. In contrast, AttUCT achieved high-quality artifact-free SOS reconstruction with sharp anatomical delineation and accurate SOS values across both soft and bone tissues, demonstrating high concordance with ground-truth maps.

### 2) Comparison with Deep Learning Methods

The qualitative comparison of reconstruction results on the test dataset is presented in Fig. 5, where AttUCT exhibited superior imaging quality compared to other deep learning methods, particularly in visualizing fine anatomical structures such as blood vessels and small tendons (highlighted in yellow boxes). The quantitative analysis in Table 2 further confirmed this advantage, demonstrating that AttUCT achieved the best reconstruction accuracy with the highest PSNR (29.23 dB) and SSIM (0.928), as well as the best perceptual similarity with the lowest LPIPS (0.042) among all deep learning methods.

### 3) Ablation Study Analysis

To validate the contribution of each attention component, we conducted ablation experiments with four variants (Table 2). The AttUCT (baseline) removed all attention mechanisms. AttUCT (CA) incorporated channel attention (CA) in the signal input head while removing spatial attention from the backbone. AttUCT (SA) integrated spatial attention (SA) mechanisms (attention gates and self-attention module) in the U-Net backbone but excluded channel attention. The full AttUCT (CA+SA) combined both channel and spatial attention components, achieving optimal reconstruction performance and demonstrating the synergistic effectiveness of the complete attention-enhanced architecture.

Table 2. Quantitative Comparison of Deep Learning Methods on Simulated Dataset

| Method | PSNR | SSIM | LPIPS |
|---|---|---|---|
| DEUNet-FE[24] | 25.71 | 0.901 | 0.067 |
| ViT[33] | 25.99 | 0.897 | 0.064 |
| DeepUCT[23] | 27.17 | 0.913 | 0.063 |
| AutoSOS[22] | 27.40 | 0.913 | 0.056 |
| AttUCT (baseline) | 27.85 | 0.922 | 0.049 |
| AttUCT (CA) | 28.89 | 0.926 | 0.052 |
| AttUCT (SA) | 28.81 | 0.924 | 0.053 |
| **AttUCT (CA+SA)** | **29.23** | **0.928** | **0.049** |

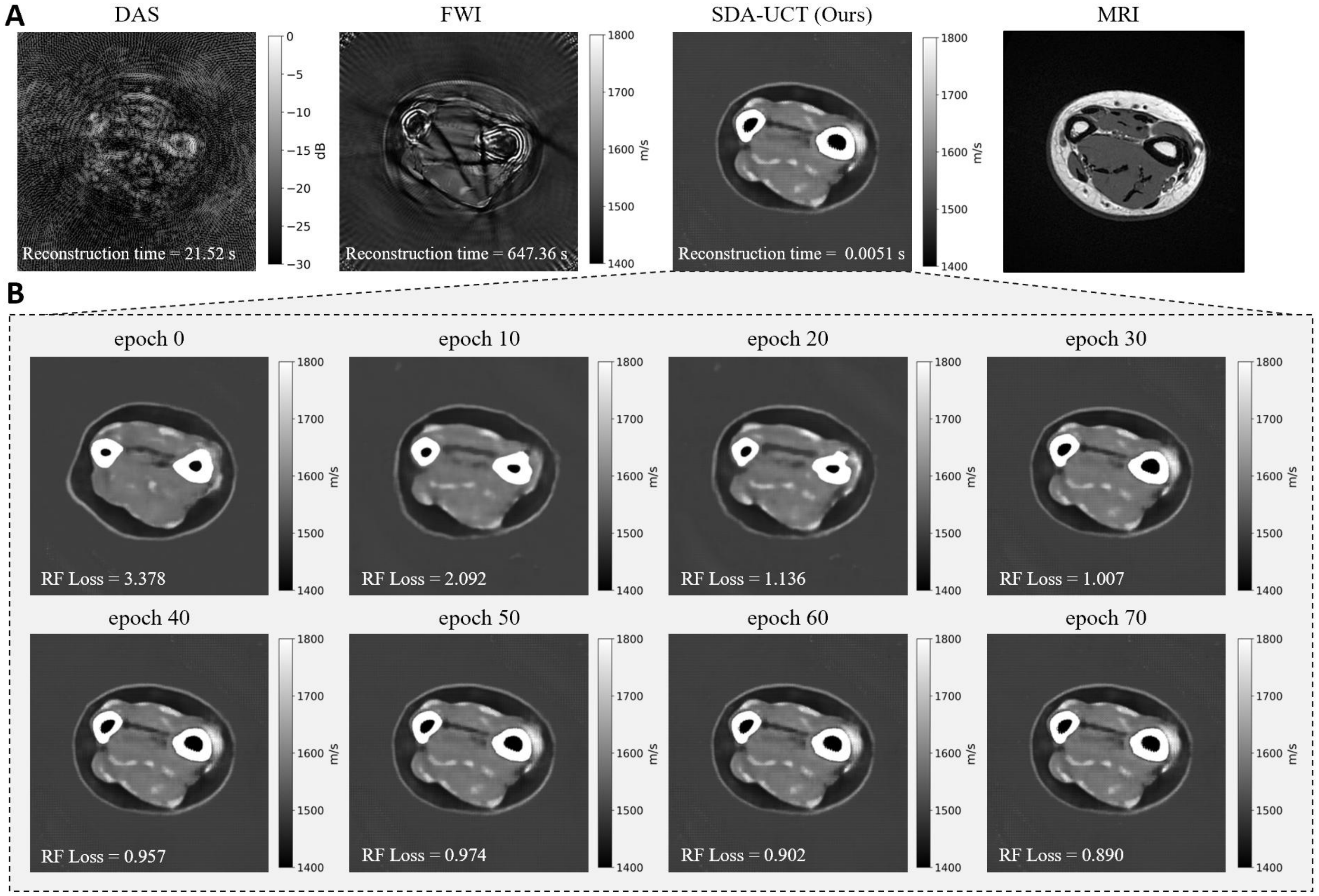

Figure 6. (A) Comparison of SDA-UCT with conventional methods on in-vivo data. (B) Evolution of SOS reconstruction during self-supervised adaptation.

### D. Self-Supervised Adaptation on in-vivo Experimental Dataset

Following Stage 1 supervised pre-training on simulation data with ground-truth SOS maps, we performed full-parameter self-supervised fine-tuning for SDA-UCT on the *in-vivo* experimental dataset without labeled SOS maps.

#### 1) Comparison with Conventional Methods

Figure 6(A) shows imaging results for SDA-UCT and conventional methods on in-vivo experimental data. DAS produced low-quality B-mode images with severe blurring and speckle noise, showing only coarse bone contours. FWI displayed improved structural information but suffered from severe artifacts in bone regions. In contrast, SDA-UCT successfully reconstructed artifact-free SOS images with clear delineation of complex anatomical structures, demonstrating high concordance with MRI references.

#### 2) Progressive Improvement for SDA-UCT During Fine-Tuning

Figure 6(B) illustrates the evolution of SDA-UCT model predictions during the fine-tuning process. Before fine-tuning (epoch 0), the SDA-UCT predictions exhibit pronounced structural abnormalities with little detailed information, reflecting the domain gap between simulation-trained models and real in-vivo experimental data. As self-supervised fine-tuning progresses, the SDA-UCT gradually adapted to the feature distribution of in-vivo experimental data and continuously improved reconstruction performance with gradually decreased RF loss. At convergence (epoch 70), the SDA-UCT accurately reconstructed anatomical structures consistent with MRI images, demonstrating satisfactory domain adaptation. The results indicate that supervised pre-training with SOS loss provides essential initialization while self-supervised fine-tuning with RF loss effectively bridges the simulation-to-real domain gap.

#### 3) Learning Rate Optimization for Fine-Tuning

To determine the optimal learning rate for self-supervised fine-tuning, we conducted a systematic evaluation across six learning rates ranging from 1e-7 to 1e-2 (Table 4). Each configuration was trained for 100 epochs on experimental data, with the final loss values summarized in Table 4. The results demonstrate that a learning rate of 1e-4 achieves the lowest RF loss.

Table 4. SDA-UCT RF Loss Under Different Learning Rates

| Learning rare | Loss value after fine-tuning |
|---|---|
| 1e-2 | 2.528 |
| 1e-3 | 2.482 |
| 1e-4 | 0.715 |
| 1e-5 | 0.832 |
| 1e-6 | 1.045 |
| 1e-7 | 1.771 |

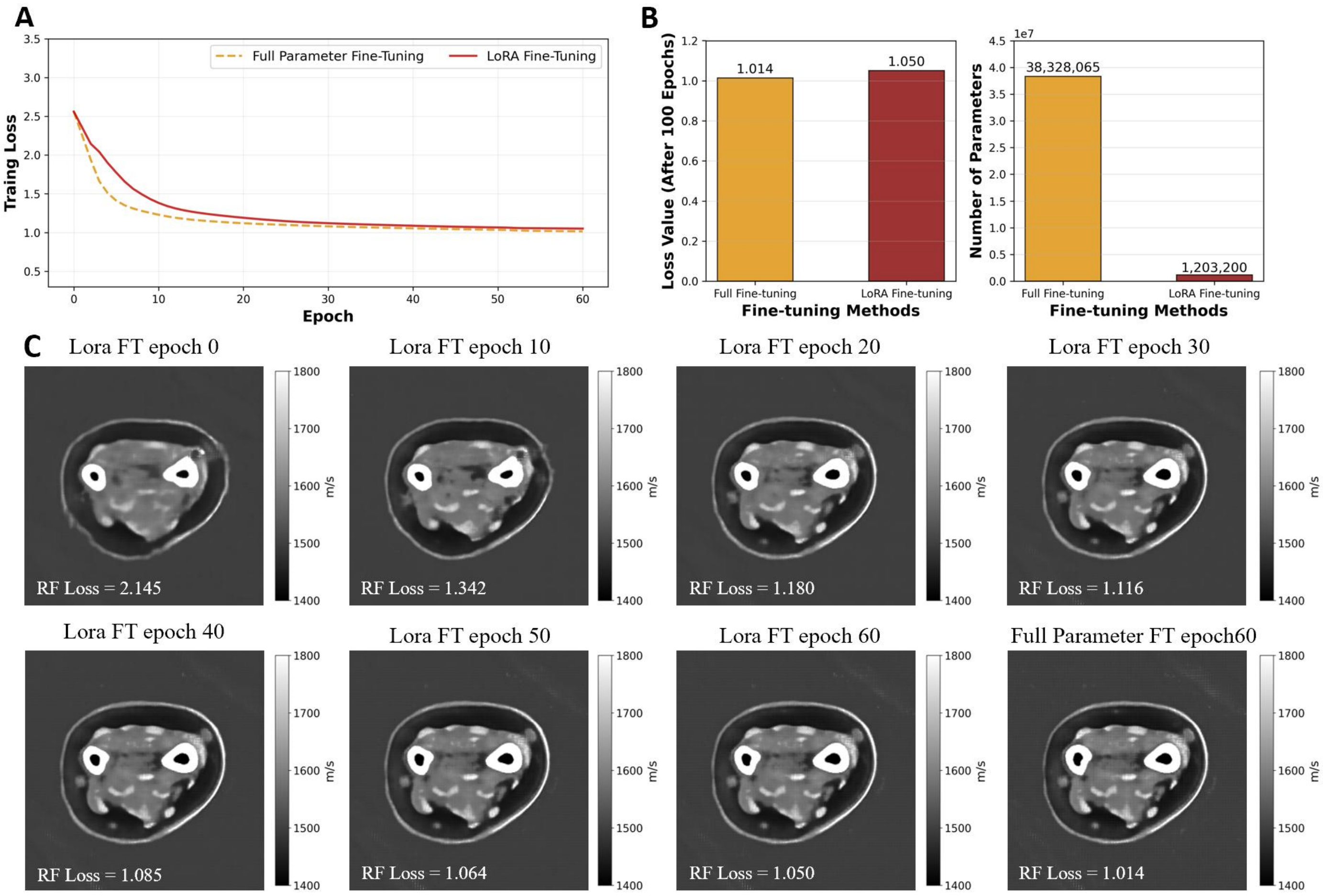

Figure 7. (A) Loss convergence curves for LoRA and full fine-tuning. (B) Parameter efficiency and final loss comparison. (C) Evolution of SOS reconstruction during LoRA fine-tuning.

### *E. LoRA-Based Sample-Specific Adaptation*

To address potential reconstruction degradation for samples with atypical acoustic properties, we employed LoRA-based fine-tuning for efficient patient-specific adaptation. This mechanism adjusts only 3% of model parameters while maintaining comparable performance to full fine-tuning. The LoRA experiments were conducted on one specific volunteer (20 samples).

#### *1) Progressive Improvement During LoRA Fine-Tuning*

Figure 7(C) illustrates the progressive improvement during LoRA fine-tuning. As training progressed, RF loss significantly decreased and reconstruction quality continuously improved. Initial reconstructions exhibited blurred boundaries, while continued training progressively revealed clearer anatomical features, particularly for bone, vascular, and tendon structures.

#### *2) Fine-tuning Efficacy and Parameter Efficiency Analysis*

Figure 7(A) compares training loss curves between LoRA and full-parameter fine-tuning. Both approaches exhibited satisfactory convergence and achieved comparable final RF loss values, confirming that LoRA preserves fine-tuning effectiveness while substantially reducing computational complexity.

Figure 7(B) shows the parameter efficiency of LoRA. With rank $r = 16$, LoRA achieved reconstruction accuracy comparable to full-parameter fine-tuning while utilizing only 3% of the parameters (1.2M vs. 38.3M). This substantial parameter reduction demonstrates the clinical feasibility of patient-specific fine-tuning in resource-constrained scenarios.

### *F. Fast and Three-Dimensional UCT Reconstruction for Forearm*

To evaluate the three-dimensional reconstruction performance for SDA-UCT, we performed layer-by-layer SOS reconstruction on 20 consecutive cross-sectional scans axially along the human forearm. The SDA-UCT reconstructed each cross-sectional SOS image in 5 ms, yielding a total reconstruction time of 100 ms for the complete 20-layer 3D volume reconstruction. The reconstruction time for FWI is 647.362 seconds for one cross-sectional SOS image. The SDA-UCT achieved approximately a five-order-of-magnitude improvement (0.005 s for SDA-UCT vs 647.362 s for FWI) over traditional FWI methods for comparable UCT tasks. Figure 8 presents the three-dimensional SOS reconstruction for human forearm using volume-rendered using 3D Slicer software, where adjusting SOS thresholds enables selective visualization of different tissues.

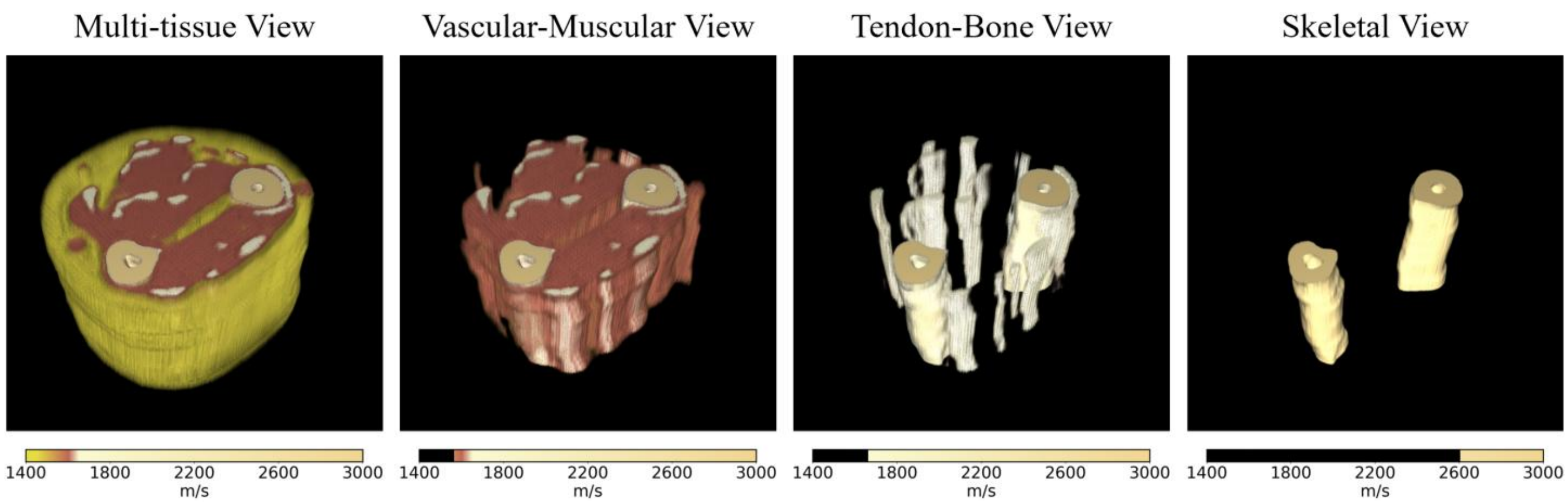

Figure 8. Three-dimensional tissue visualization using SDA-UCT with different SOS thresholds.

## IV. DISCUSSION

Conventional FWI methods iteratively reconstructing SOS via locally gradient-descent optimization have showed promising results for soft tissues (such as the breast) imaging but struggle significantly to image musculoskeletal tissues with large acoustic impedance contrasts. In this study, the proposed SDA-UCT successfully reconstructed high-quality SOS images showing complex anatomical structures (including the soft tissues and bones) for human forearm, in high concordance with the MRI references. The key factors for SDA-UCT include effectively bridging the simulation-to-real domain gap via the two-stage domain-adaptative framework with the supervised pre-training on large-scale simulation datasets and the self-supervised fine-tuning on *in-vivo* experimental datasets. Specifically, the AttUCT was designed for the task of UCT imaging musculoskeletal tissues with large SOS discrepancy, followed by a physics-informed self-supervised learning framework without the need for *in-vivo* ground-truth SOS labels. Providing high-quality imaging with little artifacts, the SDA-UCT may help to extend the clinical application of UCT in musculoskeletal disease diagnosis, including tendon tears, ligament injuries, fracture assessment, and early-stage arthritis detection.

The fast reconstruction time (i.e., only 5 ms per frame) for SDA-UCT represents a five-order-of-magnitude acceleration over conventional FWI methods. The millisecond-level reconstruction speed enables continuous 2D imaging at a frame rate exceeding 200 Hz, which is sufficient for capturing physiological motion and tissue deformation during musculoskeletal functional assessments. Furthermore, the SDA-UCT supports near-real-time 3D volumetric reconstruction by sequentially processing multiple 2D slices, enabling comprehensive anatomical visualization within seconds. The 3D volumetric reconstruction provides standard visualization for biological tissues with better interpretability and help to reduce the operator- dependent reliance on B-mode ultrasound imaging.

The integrated LoRA mechanism utilizing only 3% of model parameters achieves comparable performance to full fine-tuning, enabling patient-specific model customization with minimal storage overhead. This LoRA fine-tuning approach effectively addresses the long-tail distribution problem [40], facilitating personalized imaging for atypical anatomical variations in precision medicine applications.

Despite achieving fast and accurate musculoskeletal imaging, SDA-UCT has several limitations. First, the limited dataset scale (1,000 simulated and 160 experimental samples) constrains model generalization capabilities. Second, since the validation was restricted to the dataset from human forearms, the SDA-UCT effectiveness on other musculoskeletal part (such as the lower extremities and cranial tissues) remains unverified. Third, limited atypical pathological samples do not provide a comprehensive validation of the LoRA fine-tuning mechanism.

Future work will address aforementioned limitations in several directions. First, expanding the dataset scale and diversity will enhance the model's generalization capability. Second, extending data acquisition to lower extremities and transcranial regions will validate the framework's applicability across varied clinical scenarios. Third, collecting additional atypical pathological samples will enable comprehensive validation of the LoRA personalization mechanism.

## V. CONCLUSION

This work presents SDA-UCT, the first end-to-end deep learning framework for *in-vivo* musculoskeletal ultrasound computed tomography through physics-informed self-supervised domain adaptation. By integrating supervised pre-training on simulation data with self-supervised fine-tuning guided by wave equation constraints, the SDA-UCT addresses the *in-vivo* ground-truth label scarcity, and achieved a rapid reconstruction time (5 ms per frame) and high-resolution (29.23 dB PSNR and 0.928 SSIM on labeled data) SOS imaging for human forearm *in vivo*. The 3D volumetric reconstruction provides standard visualization with better interpretability for musculoskeletal structures, showing potential for disease diagnosis in biological tissues *in vivo*.